\documentclass[letterpaper, 10 pt, conference]{ieeeconf}  
\IEEEoverridecommandlockouts                              
\usepackage{graphicx}
\usepackage{cite}
\usepackage{picinpar}
\usepackage{url}
\usepackage[latin1]{inputenc}
\usepackage{colortbl}
\usepackage{soul}
\soulregister\cite7
\soulregister\ref7
\usepackage{multirow}
\usepackage{pifont}
\usepackage{color}
\usepackage{alltt}
\usepackage[hidelinks]{hyperref}
\usepackage{enumerate}
\usepackage{siunitx}
\usepackage{breakurl}
\usepackage{epstopdf}
\usepackage{pbox}
\usepackage{amsmath,amssymb,amsfonts}
\usepackage{algorithmic}
\usepackage{textcomp}
\usepackage{wrapfig}
\usepackage{empheq}
\usepackage{array}
\usepackage{makecell}
\usepackage{xcolor}
\usepackage{booktabs}
\usepackage{threeparttable}
\usepackage{adjustbox}
\usepackage{subfigure}
\usepackage{verbatim}
\usepackage[misc]{ifsym}
\graphicspath{{figures/}}
\usepackage{hyperref}

\title{\LARGE \bf
Multi-Stage Reinforcement Learning for Non-Prehensile Manipulation
}

\author{Dexin Wang, Faliang Chang, Chunsheng Liu %
\thanks{*This work was supported in part by National Natural Science Foundation of China (NO. U22A2058, 62176138, 62176136), National Key R\&D Program of China (NO.2018YFB1305300), Shandong Provincial Key Research and Development Program (NO. 2019JZZY010130, 2020CXGC010207).}
\thanks{All authors are with the School of Control Science and Engineering, Shandong University, Ji'nan, Shandong 250061, China.}%
}

\begin{document}

\maketitle
\thispagestyle{empty}
\pagestyle{empty}

\begin{abstract}

Manipulating objects without grasping them enables more complex tasks, known as non-prehensile manipulation.
Most previous methods only learn one manipulation skill, such as reach or push, and cannot achieve flexible object manipulation.
In this work, we introduce MRLM, a Multi-stage Reinforcement Learning approach for non-prehensile Manipulation of objects.
MRLM divides the task into multiple stages according to the switching of object poses and contact points.
At each stage, the policy takes the point cloud-based state-goal fusion representation as input, and proposes a spatially-continuous action that including the motion of the parallel gripper pose and opening width.
To fully unlock the potential of MRLM, we propose a set of technical contributions including the state-goal fusion representation, spatially-reachable distance metric, and automatic buffer compaction.
We evaluate MRLM on an Occluded Grasping task which aims to grasp the object in configurations that are initially occluded. 
Compared with the baselines, the proposed technical contributions improve the success rate by at least 40\% and maximum 100\%, and avoids falling into local optimum.
Our method demonstrates strong generalization to unseen object with shapes outside the training distribution.
Moreover, MRLM can be transferred to real world with zero-shot transfer, achieving a 95\% success rate.
Code and videos can be found at \href{https://sites.google.com/view/mrlm}{https://sites.google.com/view/mrlm}.

\end{abstract}

\section{INTRODUCTION}

Non-prehensile skills can improve the robot's dexterity in manipulating objects.
However, the inconsistent contact between the robot and object introduces a changing dynamic model, which increases the difficulty of manipulation.
Most previous methods only learn one manipulation skill, such as reach or push, and thus are applicable only to simple tasks.

In order to complete more flexible tasks, some researchers plan the entire manipulation process in advance based on random-sampling \cite{pollayil2021planning, jiang2022path}, search \cite{liang2022search, garrett2015backward} and optimization \cite{junge2020improving, stouraitis2020online}.
However, planning the entire process is time-consuming and struggles with complex-shaped objects.
Some researchers use reinforcement learning algorithms to train agents to perform tasks, which improves planning efficiency, but there are still limitations.
Sun \textit{et al.} limit robotic motion to tilting objects \cite{sun2020learning}.
Zhou \textit{et al.} represent the object as a minimum circumscribed box, which cannot handle objects with complex shapes \cite{zhou2023learning1}.
Zhou \textit{et al.} train the agent end-to-end to complete long-horizon tasks, but the sample efficiency is low due to sparse rewards \cite{zhou2023learning2}.
In contrast, we take the desired 6-DOF object pose as task goal, allowing the robot to explore arbitrary skills to accomplish the task.
We use point clouds to represent objects to handle different shapes.
Moreover, we decomposes tasks into multiple stages whose goals are easier to explore, thereby improving sample efficiency.

\begin{figure}[tp]
	\centering
	{\includegraphics[scale=0.5]{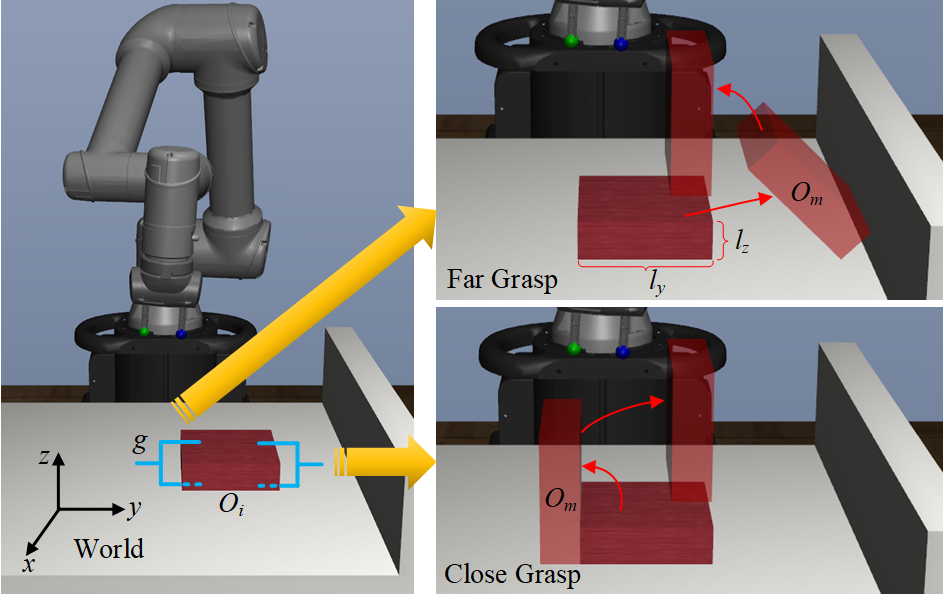}}
	\caption{
		The process of occluded grasping task.
		Given a goal object pose and predefined grasp pose $g$, our proposed MRLM calculates the corresponding intermediate object pose $O_m$, and predicts the pose change and opening width change of the gripper, so that the object moves between poses and finally reaches the goal pose.
	}
	\label{fig_inter_pose}
\end{figure}

\begin{figure*}[tp]
	\centering
	{\includegraphics[scale=0.8]{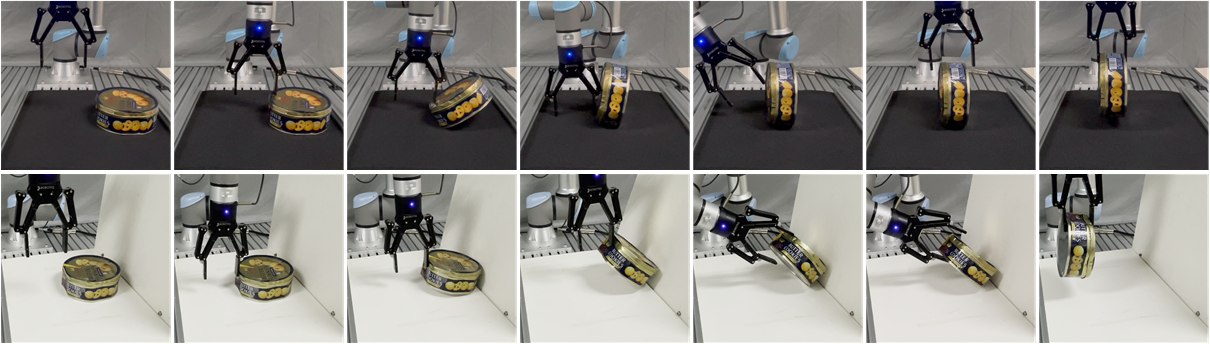}}
	\caption{
		We present a system for non-prehensile manipulation and apply it to the ``Occluded Grasping" task. 
		The goal is to grasp the object for which all grasp configurations are initially blocked.
		The figure shows two examples of the emergent behaviors of MRLM.
	}
	\label{fig_example}
\end{figure*}

In this work, we introduce MRLM, a Multi-Stage Reinforcement Learning approach for non-prehensile Manipulation of objects. 
Given the initial and goal poses of an object, MRLM first computes a set of contact points and intermediate object poses, which satisfy two conditions:
(1) The object can move between adjacent poses when the robot applies a continuous force to the contact point on the object surface;
(2) The object is in static equilibrium at all poses.
Then, MRLM divides the task into multiple operational stages according to the switching of contact points and object poses.
We use off-policy RL algorithm to train a \textbf{P}oint cloud \textbf{M}otion based \textbf{Man}ipulation \textbf{Net}work (P2ManNet) as the policy to complete each stage successively.
In each stage, the policy takes the point cloud-based state-goal fusion representation as input, and proposes a spatially-continuous action that including the motion of gripper pose and opening width.
We introduce reward shaping to improve learning efficiency, and propose a spatially-reachable distance metric to compute the reward function which avoid the policy getting stuck in local optima.
We build a buffer for each stage and automatically compact the training data in the buffer according to the learning progress, so as to accelerate the learning of the current stage without forgetting the skills of previous stages.

We apply MRLM to an occluded grasping task which aims to grasp the object in configurations that are initially occluded (Fig.~\ref{fig_example}).
Compared with the baselines, the state-goal fusion representation, spatially-reachable distance metric, and automatic buffer compaction improve the success rate by an average of 100\%, 40\%, and 60\%, respectively.
Compared with single-stage RL method, MRLM improves the success rate by 100\%.
MRLM demonstrates strong generalization to unseen object with shapes outside the training distribution.
Moreover, MRLM can be transferred to real world with zero-shot transfer, achieving a 95\% success rate.

\section{RELATED WORK}
\textbf{Non-prehensile manipulation}. Non-prehensile manipulation allows robots to manipulate objects using skills other than grasping.
Manipulation tasks in real scenes often involve multiple changes in the object's contact with the robot and the scene, leading to inconsistent dynamics models and introducing significant challenges for scene description, planning, and control \cite{kroemer2021review}.
Recent works mainly apply sampling-based and learning-based methods, but most of them are limited to single or combinatorial skills \cite{gao2022fast, shridhar2022cliport} and primitive-shaped objects \cite{cheng2022contact, yang2021hierarchical}.
In contrast, our work shows more sophisticated skills at coping with changing dynamics models while generalizing to object geometries outside the training distribution.

\textbf{Manipulation with point clouds}.
Manipulating objects based on point clouds has been studied for decades \cite{jang2005visibility, rusu2009close}.
Early works mainly use deep learning techniques to learn state estimation \cite{hoang2022voting, wang2022six} or manipulation skills from point clouds \cite{tang2022track, huang2022task}.
As manipulation tasks become complex and labels are difficult to obtain, reinforcement learning techniques are gradually introduced to learn more dexterous skills from point clouds \cite{lu2022excavation, you2021omnihang}.
However, most works employ separate networks to encode point cloud states and task goals, resulting in inconsistent feature spaces, making it difficult for the policy to converge \cite{qin2023dexpoint, liu2022frame}.
In contrast, we use point clouds to encode state and task goals simultaneously, keeping the feature space consistent.


\section{PROBLEM DESCRIPTION}
We focus on the task of occluded grasping with non-prehensile manipulation.
Occluded grasping is a special class of grasping tasks where the required grasp pose is initially blocked by a table, wall or other object.
The robot has to interact with the object to make the grasp pose reachable.

The detailed task settings and assumptions are as follows:
\begin{enumerate}[1)]
	\item
	\textbf{Scene}:
	The scene contains a fixed table and wall, as well as a object.
	We focus flat objects because they cannot be grasped without reorientation.
	The models of all parts in the scene are known.
	Before the task starts, the object is on the table and all grasping configurations are blocked.
	
	\item 
	\textbf{Object goal pose}: 
	Located above the table, as shown by the uppermost translucent object in Fig.~\ref{fig_inter_pose}.
	
	\item 
	\textbf{Grasp pose}: 
	Grasp pose $g$ is given before the task starts.
	We assume that the grasp pose is on the side of the object surface close to or away from the wall (Fig.~\ref{fig_inter_pose}), which represents the two extremes of all grasps.
	We use \textit{close grasp} and \textit{far grasp} to denote the grasp pose on the side of the object surface close to and away from the wall, respectively.
	
	\item 
	\textbf{Intermediate object poses}:
	For different grasp poses, we assume that there is only one intermediate object pose $O_m$.
	The intermediate object pose corresponding to each grasp pose is shown in Fig.~\ref{fig_inter_pose}.
	It should be emphasized that $O_m$ is not constant, but is calculated based on the object initial pose, which is detailed in Section.~\ref{task_decomposition}.
	
\end{enumerate}

\section{METHOD}

\subsection{Task Decomposition}
\label{task_decomposition}
We decompose the task based on the switching of object poses and contact points between the gripper and object.
For different grasp poses, we calculate the corresponding intermediate object pose $O_m$ and contact point respectively.

$O_m$ is calculated based on the static equilibrium principle, as shown in Fig.~\ref{fig_inter_pose}.
If $g$ is on the side of the object surface close to the wall, $O_m$ is obtained by rotating the object by $90$ degrees along the x-axis of the world frame.
The object is in static equilibrium under its own gravity and the support force of the table.
$O_m$ is calculated by $O_i * T_{O_i}^{O_m}$, where $O_i$ denotes object initial pose and $T_{O_i}^{O_m}$ denotes the transformation from $O_i$ to $O_m$:
\begin{equation}
	T_{O_i}^{O_m} = 
	\begin{bmatrix}
		1	& 0 	& 0 	& 0	\\
		0 	& 0 	& -1 	& -l_y-l_z 	\\
		0   & 1 	& 0  	& l_y-l_z	\\
		0 	& 0 	& 0		& 1
	\end{bmatrix}
	\label{eq_1}
\end{equation}
where $l_y$ and $l_z$ represent the dimensions of the smallest circumscribed box of the object on x and y axes, respectively.

If $g$ is on the side of the object surface away from the wall, $O_m$ is obtained by translating the object along the y-axis and rotating it by $-45$ degrees along the x-axis of the world frame, while the object is leaning against the wall.
The object is in static equilibrium under its own gravity, the support force of the wall and table, and the contact force from the gripper.
$O_m$ is represented as:
\begin{equation}
	O_m = 
	\begin{bmatrix}
		1	& 0 	& 0 	& 0	\\
		0 	& \frac{\sqrt{2}}{2}	 	& \frac{\sqrt{2}}{2} 	& W-\frac{l_y+l_z}{\sqrt{2}} 	\\
		0   & -\frac{\sqrt{2}}{2} 	& \frac{\sqrt{2}}{2}  	& T+\frac{l_y+l_z}{\sqrt{2}}	\\
		0 	& 0 	& 0		& 1
	\end{bmatrix}
	\label{eq_2}
\end{equation}
where $W$ denotes the y-coordinate of wall and $T$ denotes the z-coordinate of table.

\begin{figure}[tp]
	\centering
	{\includegraphics[scale=0.38]{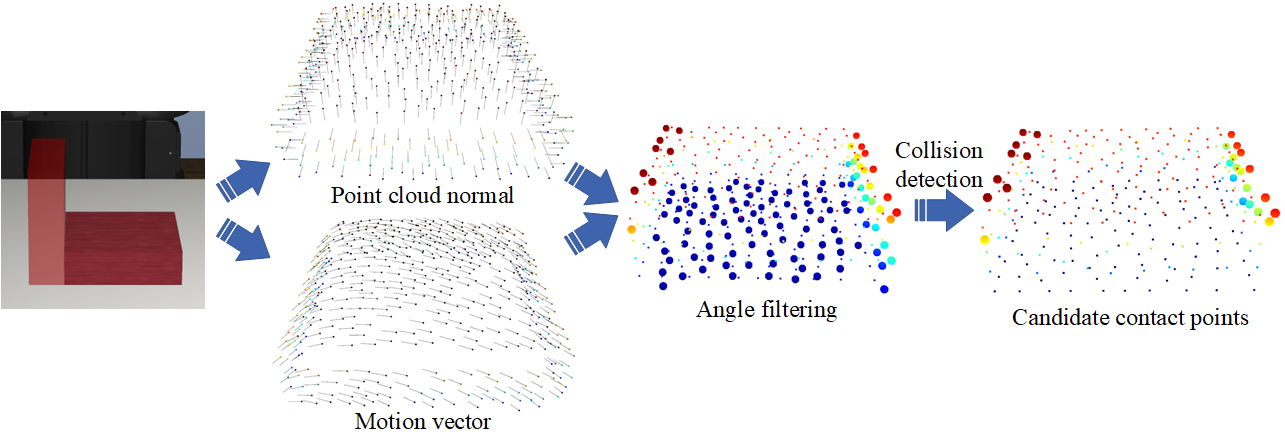}}
	\caption{
		Candidate contact points calculation pipeline.
		Firstly, the point cloud is sampled and the normal and motion vectors are calculated, then the points whose vector angle is greater than the threshold are filtered, and finally the candidate contact points without collision with the scene are reserved.
	}
	\label{fig_contact_pt}
\end{figure}

The contact point is calculated based on the following principle: the object can move between adjacent poses when the robot applies a continuous force to the contact point.
First, MRLM uniformly samples a set of points on the object surface and calculates the normal towards the interior, and then calculates the angle between the normal and the motion vector of each point.
Motion vector is represented as:
\begin{equation}
	v = p_m - p_i
	\label{eq_3}
\end{equation}
where $p_m$ and $p_i$ are the coordinates of the point in world frame when the object pose is $O_m$ and $O_i$, respectively.
Points whose angle is greater than the threshold are filtered.
Finally we compute candidate contact points satisfying the following conditions:
(1) located in the opposite direction of the normal of the sampling points,
(2) the distance from the object surface is equal to the radius of the gripper finger, and (3) have no collision with the scene, as shown in Fig.~\ref{fig_contact_pt}.

During training, we employ RL algorithm to explore how to accomplish the task (i.e., action) and the quality of manipulation (i.e., Q value) based on candidate contact points.
During testing, the contact point with the largest Q value is selected for manipulation.
Based on different object poses and contact points, 
We decompose the task into four stages:
\begin{enumerate}[1)]
	\item
	\textbf{Stage 1}:
	The gripper moves to the contact point $p$, and the object remains in the initial pose $O_i$.
	
	\item
	\textbf{Stage 2}:
	The gripper remains in contact with the object at $p$, and the object moves to $O_m$.
	
	\item
	\textbf{Stage 3}:
	The gripper moves to the contact points $p_g$ corresponding to the grasp pose $g$, and the object remains in $O_m$.
	
	\item
	\textbf{Stage 4}:
	The gripper is closed and move the object to the goal pose.
\end{enumerate}

Note that $p$ is the goal contact point of one finger of the gripper with the object in the first two stages (i.e., the blue finger in Fig.~\ref{fig_inter_pose}), and the goal for the other finger is set to a position where the spread gripper does not collide with the scene and has the least x-axis rotation.
The robot completes the first three stages by executing the action output by the policy, and completes the fourth stage through the predefined program.
A stage is defined to be successful if the following conditions are met at the end of an episode:
\begin{enumerate}[1)]
	\item
	The position difference and the orientation difference of the object pose and goal pose are less than the pre-defined thresholds $\epsilon_d$ and $\epsilon_\theta$.
	\item
	The position difference of the gripper fingers and the target contact points is less than $\epsilon_d$.
\end{enumerate}

\subsection{State-Goal Representation}
\begin{figure*}[tp]
	\centering
	{\includegraphics[scale=0.47]{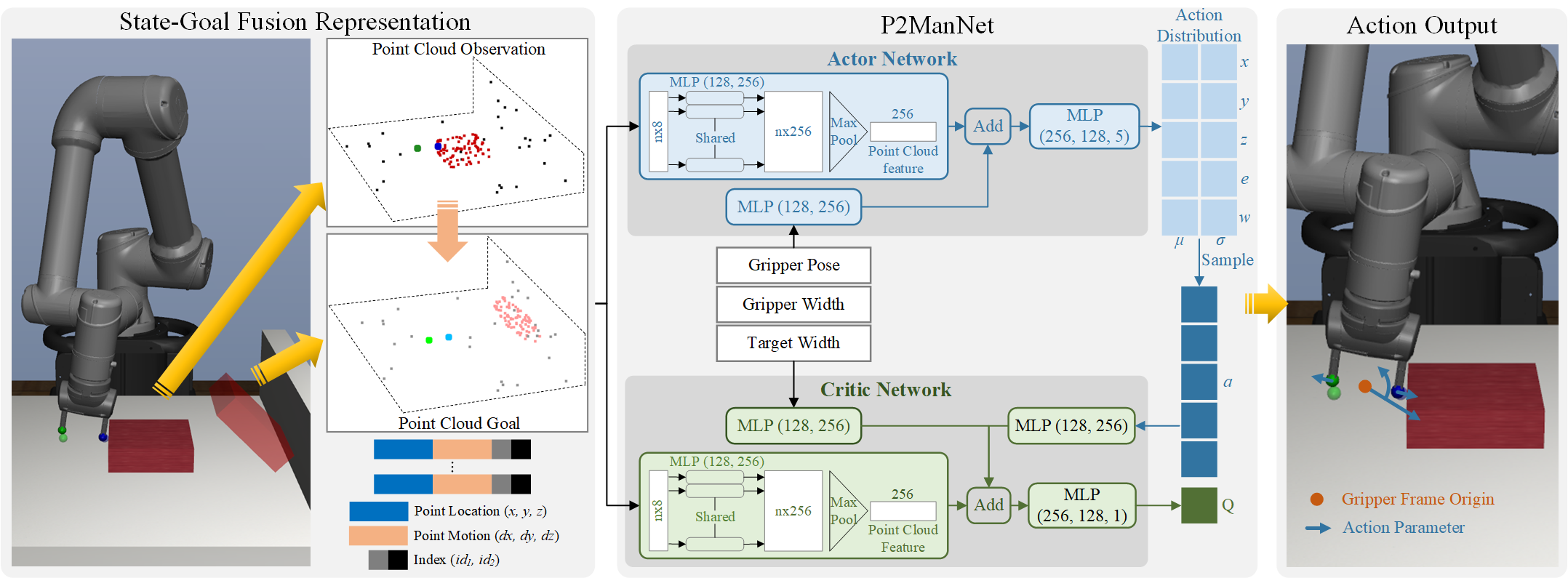}}
	\caption{
		Overview of our network.
		The state-goal fusion representation includes the location, motion and index of points of the object, gripper finger and background.
		The point cloud goal of the finger is obtained based on the sampled candidate contact points.
		The motion is the location difference between the point cloud observation and the goal.
		P2ManNet takes observation and goal as input and outputs action distribution and Q value.
		Semi-transparent models represent the goal location of object and fingers.
	}
	\label{fig_net}
\end{figure*}

We propose a point cloud-based state-goal representation that simultaneously represents observation states and task goals.
As shown in Fig.~\ref{fig_net}, the point cloud consists of four parts: object, gripper fingers, tabletop, and wall.
The point cloud of the object is exported based on the 3D model.
The point clouds of the tabletop and wall only contain two planes that are perpendicular to each other.
We designed the gripper fingers to be spherical, so that when the finger applies force to the contact point from any direction, the distance between the object surface and the finger center is same.

The point cloud includes the location $(x, y, z)$, motion $(dx, dy, dz)$ and index $(id_1, id_2)$ of each point.
Locations represent real-time observations.
Motions are the difference between the target coordinates of points and the current coordinates, which represent the goal.
Both locations and motions are in the world frame.
The table and wall make up the motionless background, indexed at $(0, 0)$.
The indices of the two fingers of the gripper are $(0, 1)$ and $(1, 0)$ respectively.
The index of the object is $(1, 1)$.
In our experiments, the number of point clouds for background, gripper fingers and object are $\{30, 228, 100\}$, respectively.

\begin{figure}[tp]
	\centering
	{\includegraphics[scale=0.7]{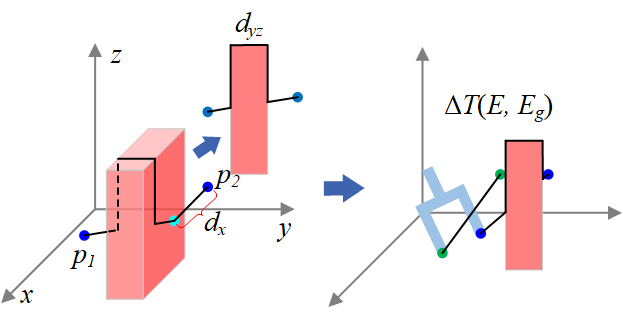}}
	\caption{
		Spatially-reachable distance metric.
		The distance between points $p_1$ and $p_2$ is equal to the sum of the shortest reachable distance between projected points on the yz plane $d_{yz}$ and the x coordinate difference $d_x$.
		The gripper translational distance $\Delta T(E, E_g)$ is equal to the sum of the spatially-reachable distances of the two fingers.
	}
	\label{fig_distance}
\end{figure}

\subsection{RL Algorithm}
\textbf{Observation and Action Space}:
In addition to the points location in the state-goal representation, the observation that is input to the policy also includes gripper pose in the world frame, gripper opening width and target width.
The action space of the policy is five dimensional:
\begin{equation}
	a = (t, e, w)
	\label{eq_action}
\end{equation}
where $t \in \mathbb{R}^3$ and $e \in \mathbb{R}$ indicate the desired change in the position and x-axis rotation of the gripper, with a maximum of 1 $cm$ and 5 degrees, respectively. 
$w \in \mathbb{R}$ indicates the desired change in the gripper opening width, with a maximum of 6 $mm$.
Actions are executed through a lower-level Operation Space Controller (OSC).
An outline of the policy execution is shown in Fig.~\ref{fig_net}.

\textbf{Network}:
In order to predict spatially continuous actions, we design a \textbf{P}oint cloud \textbf{M}otion based \textbf{Man}ipulation \textbf{Net}work (\textbf{P2ManNet}) as the policy, as shown in Fig.~\ref{fig_net}.
P2ManNet includes an actor network and a critic network, both of which contain an encoder with the same structure.
The encoder uses improved PointNet \cite{qi2017pointnet} and MLP to extract features from observation and goal, and then fuse them.
We remove the alignment networks in PointNet to improve efficiency.
The actor network decodes the features and predicts the action distribution, which is defined as a set of Gaussian mean $\mu$ and standard deviation $\sigma$.
Actions are sampled from the predicted distribution in training and set to the mean in testing.
The critic network fuses observation features and action features to predict the Q value.

\textbf{Reward}:
We design the same reward function for all stages based on reward shaping:
\begin{equation}
	r = - D(O, O_g) - \beta \Delta T(E, E_g) - P(E_g)
	\label{eq_r1}
\end{equation}
\begin{equation}
	D(O, O_g) = \alpha_1 \Delta T(O, O_g) + \alpha_2 \Delta \theta(O, O_g)
	\label{eq_r2}
\end{equation}
where $\alpha_1$, $\alpha_2$ and $\beta$ are the weights for the reward terms.
The first term of Equation \ref{eq_r1}, $D(O, O_g)$, is the pose difference between the target object pose $O_g$ and the current pose $O$, which is to optimize for moving the object.
This term is expanded in  Equation \ref{eq_r2} to include the translational and rotational distance.
The second term of Equation \ref{eq_r1}, $\Delta T(E, E_g)$, is the translational distance between the target gripper fingers position $E_g$ and the current position $E$, which is to optimize the contact between the gripper and object.
The third term of Equation \ref{eq_r1}, $P(E_g)$, is the occlusion penalty for the target gripper ends position, which is effective if $E_g$ is blocked by the table or wall.
All terms except penalty are scaled to the range $[0,1]$ using the \textit{tanh} function.

We design a new spatially-reachable distance metric to represent $\Delta (E, E_g)$, as shown in Fig.~\ref{fig_distance}.
Unlike Euclidean distance, the new distance metric computes the length of the shortest reachable path between two points.
Since most path searching methods are time-consuming \cite{yang2016survey}, we compute the shortest paths in $x$ and $yz$ dimensions separately and sum to approximate the shortest path in 3D space.
On the yz-dimensional plane, the shortest path $d_{yz}$ consists of truncated line between two points and rectangular outlines.
In the x-dimension, the shortest path $d_x$ is equal to the difference in the x-coordinates of two points.
With close grasp, the spatially-reachable distance metric can effectively prevent the policy from falling into local optima where the gripper reaches close to the contact points corresponding to the grasp from the side of the object instead of moving above the object to reach.

\textbf{Replay Buffer}:
Since each stage can only be explored after the previous stage is completed, a large number of trajectories of already learned stages will be accumulated in the buffer, resulting in low learning efficiency in the current stage.
To improve learning efficiency, we build a buffer for each stage and allow automatic compaction.
When the test success rate of a stage exceeds the threshold (90\% in our experiments), MRLM takes 50\% and 70\% of the data volume in the corresponding buffer at this time as the \textit{floor} and \textit{ceiling}.
When the amount of data in the buffer exceeds the \textit{ceiling}, MRLM evenly samples and deletes part of the data, so that the remaining data amount is the \textit{floor}.

\subsection{Improving Generalization}
To improve generalization across environment variations, we train the policy with Automatic Domain Randomization (ADR) \cite{akkaya2019solving}.
The policy is first trained in an initial environment with fixed parameters, and then we gradually expand the sampling boundary of environment parameters according to the test performance, such as object size, density, and friction coefficient.
Compared to directly training the policy with the largest range of parameters, ADR can reduce the need to manually adjust the parameter range and improve the convergence speed.
Table \ref{tab_sim_params} summarized the simulation parameters in the experiment, where $\Delta$ is the increment of each expansion of the parameter boundary.
Note that the objects used for training are simply boxes.

\begin{table}[tp]
	\caption{Environment parameters in the experiment.}
	\footnotesize 
	\begin{center}
		\begin{threeparttable}
			\begin{tabular}{l|ccc}
				\toprule[1pt]
				
				& Initial Value & $\Delta$ & Final Range \\
				
				\midrule[1pt]
				
				Object size x (m) & 0.15	& 0.015	& [0.12, 0.18]	\\ 
				Object size y (m) & 0.15	& 0.015	& [0.12, 0.18]	\\ 
				Object size z (m) & 0.045	& 0.005	& [0.035, 0.055]	\\ 
				Object position x (m) & 0	& 0.025	& [-0.05, 0.05]	\\
				Object position y (m) & 0.1	& 0.025	& [0.05, 0.15]	\\ 
				Object rotation z (deg) & 0	& 5	& [-10, 10]	\\ 
				Object density ($g/m^3$) & 100	& 25	& [50, 150]	\\ 
				Object friction & 0.5	& 0.1	& [0.3, 0.5]	\\ 
				Table friction & 0.5	& 0.1	& [0.3, 0.5]	\\ 
				Gripper friction & 2	& /	& [2, 2]	\\ 
				Wall position y (m) & 0.3	& /	& [0.3, 0.3]	\\ 
				
				\bottomrule[1pt]
			\end{tabular}
			\label{tab_sim_params}
		\end{threeparttable}
	\end{center}
\end{table} 

\begin{table}[tp]
	\caption{Hyperparameters for RL training.}
	\footnotesize 
	\begin{center}
		\begin{threeparttable}
			\begin{tabular}{lc}
				\toprule[1pt]
				
				Hyperparameters & Value \\
				
				\midrule[1pt]
				
				Optimizer & Adam\\
				Learning rate - Actor & 1e-3\\
				Learning rate - Critic & 1e-3\\
				Batch size & 256\\
				Soft target update ($\pi$) & 1e-2\\
				Replay buffer size & 1.5e6\\
				Discount factor ($\gamma$) & 0.95\\
				horizon for each stage & 100 \\
				
				\bottomrule[1pt]
			\end{tabular}
			\label{tab_rl_params}
		\end{threeparttable}
	\end{center}
\end{table}

\section{EXPERIMENTS}
\subsection{Simulation Experiment Setup}
We built the simulation environment to train and test MRLM using Robosuite \cite{robosuite2020}, an integrated library for robot simulation based on the Mujoco simulator \cite{todorov2012mujoco}.
The environment contains a UR5e robotic arm with a parallel gripper and an object placed on the table in front of it.
P2ManNet is trained with SAC.
Hyperparameters for training are include in Table \ref{tab_rl_params}.

In this section, we include the results in simulation to discuss each component of MRLM.
We then evaluate zero-shot sim2real transfer on a physical robot across different objects.
We compare all policies across 5 random seeds and two grasp poses, and plot the average performance with standard deviation across seeds.
The weights used for testing are the last ones saved during training.
We use 10 episodes for each evaluation setting.

\begin{figure}[tp]
	\centering
	{\includegraphics[scale=0.32]{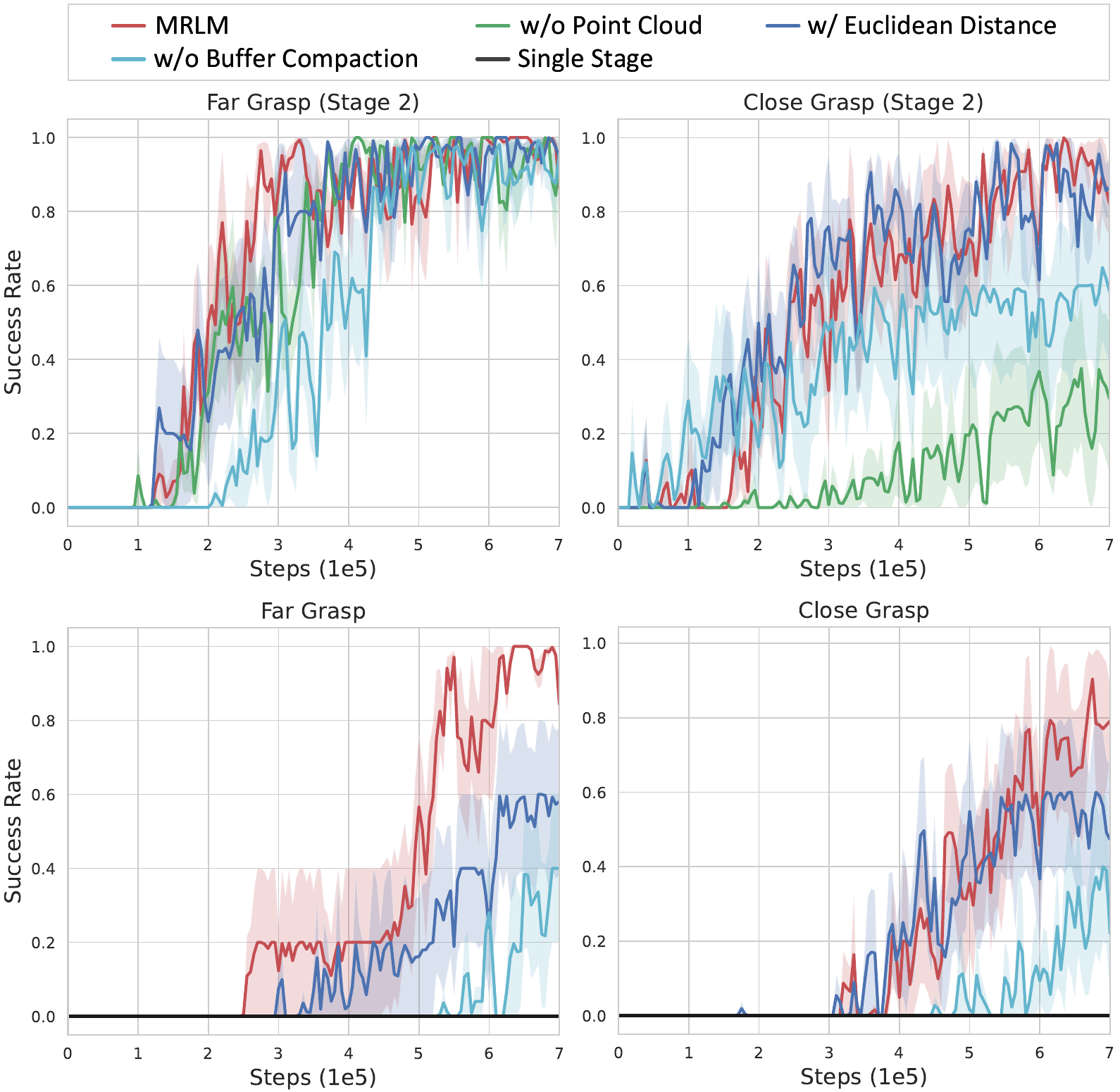}}
	\caption{
		Training curves and ablations.
	}
	\label{fig_ablation}
\end{figure}

\subsection{Ablation Experiment}
We first evaluate our method by training P2ManNet in the default environment without ADR.
Fig.~\ref{fig_ablation} shows the training curves of P2ManNet for two grasps and the ablations.
P2ManNet trained with the complete system can reach 100\% success rate with far grasp before 600,000 exploration steps and 95\% success rate with close grasp before 700,000 exploration steps.
The sample efficiency with close grasp is lower than that with far grasp due to exploring more candidate contact points.

\begin{figure}[tp]
	\centering
	{\includegraphics[scale=0.21]{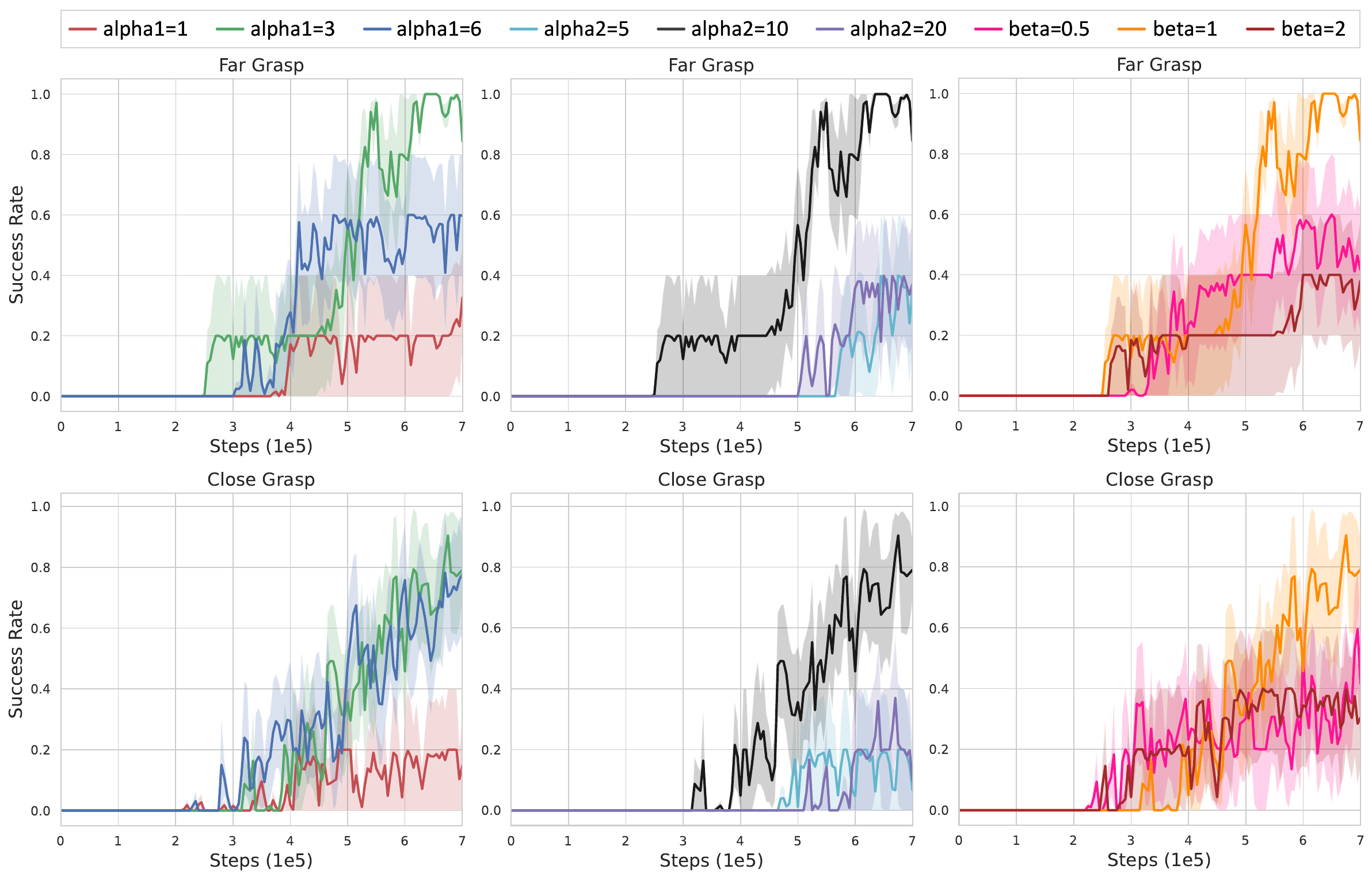}}
	\caption{Training curves with different reward weights.}
	\label{fig_reward_weights}
\end{figure}

To evaluate the importance of point cloud-based state-goal representation, we set the observation as an array containing the gripper pose, the fingers position, the object pose, the size of minimum circumscribed box, and the gripper opening width.
The goal is set as an array containing the object goal pose, the fingers goal position, and the gripper goal opening width.
The resulting policy achieves 100\% success rate on the second stage and 0\% success rate on the final goals with far grasp.
The performance with close grasp is even worse.
We speculate that the performance degradation is due to non-intuitive perception of the shape of the gripper, object, and background, making it difficult to perform precise manipulations, such as flipping and tilting.
Using the Euclidean distance instead of the proposed spatially-reachable distance to calculate the reward, MRLM has about 40\% less success rate and is more likely to get stuck at a local optima, that is, the gripper will directly approach the goal contact points causing a change in the object pose.
Without automatic buffer compaction, the sampled training data contains too many trajectories of already learned stages, resulting in low learning efficiency and a 60\% reduction in success rate..

In addition, we compare with the single-stage reinforcement learning method proposed by Zhou \textit{et al.} that accomplishes a similar occluded grasping task.
Different from our task setup, they assume that the gripper, object, and wall are in close proximity at the beginning of the task, and use the smallest circumscribed box to represent the object, so it is easier to explore the strategy of rotating the object against the wall and then accomplishing the grasp.
In our tasks with sparser rewards, the single-stage approach achieves 0\% success rate.

In all the experiments, we use $\alpha_1 = 3$, $\alpha_2 = 10$, $\beta = 1$ as the weights for the reward terms, and $P(E_g)$ is set to $5$.
To test the sensitivity of MRLM to different reward weights, we train P2ManNet with different weight values.
Fig.~\ref{fig_reward_weights} shows that MRLM is sensitive in most of the case we tested.

\begin{figure}[tp]
	\centering
	{\includegraphics[scale=0.39]{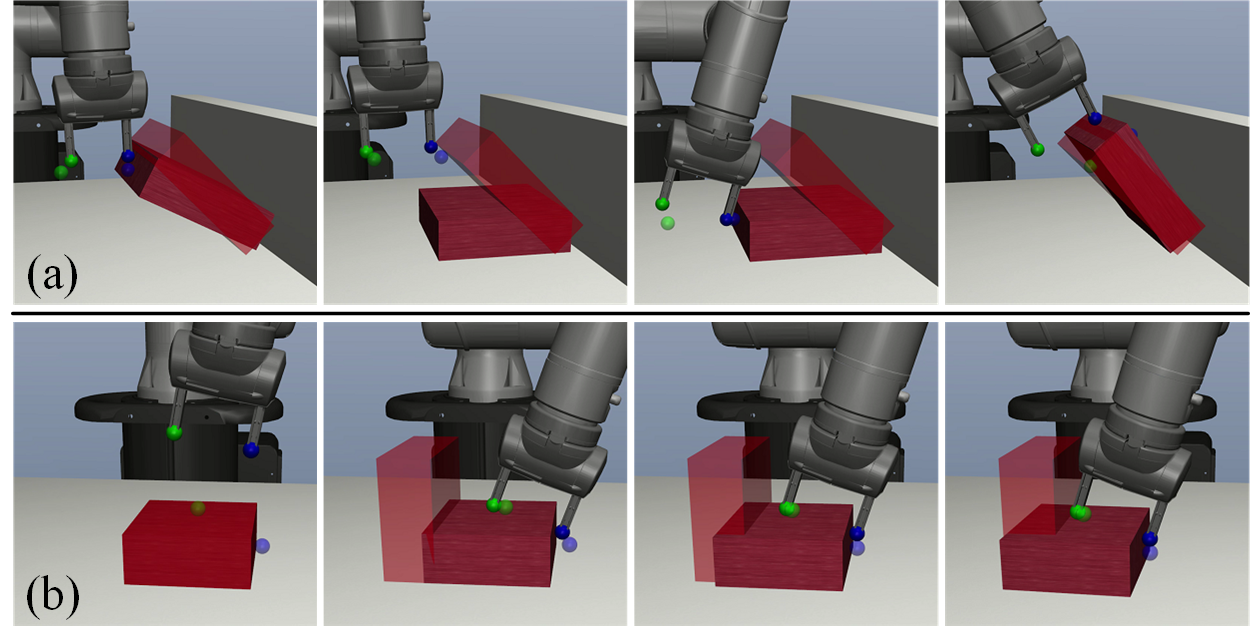}}
	\caption{
		(a) \textbf{Recover from mistakes}:
		The gripper re-tilts the object after it has fallen due to a motion mistake.
		(b) \textbf{Local optima}:
		An example of local optima with close grasp where the gripper pushes the object from the side close to the wall and then fails to flip it.
	}
	\label{fig_behaviors}
\end{figure}

\subsection{Emergent Behaviors}
Fig.~\ref{fig_example} shows typical strategies of successful policies which involve multiple stages of contact switches. 
With close grasp, the gripper first presses the contact point on the upper surface of the object on the opposite side of the grasp to flip the object, then moves above the object and then precisely approaches the contact points corresponding to the grasp.
With far grasp, the gripper first pushes the object against the wall to rotate it.
After the object reaches the intermediate pose, the gripper uses one finger to hold the object, moves another finger under the object, and then let the object drop on the bottom finger.
Finally, the gripper will try to approach the desired contact points. 
At this point, the policy has executed the grasp successfully and the object will be lifted to the goal pose.

The ability to recover from mistakes allows MRLM to overcome disturbances and reduce failure rates.
With far grasp, if the object falls while being rotated, the gripper will move to the contact point to re-rotate the object, as shown in Fig.~\ref{fig_behaviors} (a).
The same is true with close grasp, MRLM will recover from failing to flip the object.
We observe that MRLM can get stuck in local optima due to inappropriate contact points.
With close grasp, the policy may use contact points on the same side as the grasp to push the object close to the intermediate pose without successfully flipping the object (Fig.~\ref{fig_behaviors} (b)).

\begin{figure}[tp]
	\centering
	{\includegraphics[scale=0.21]{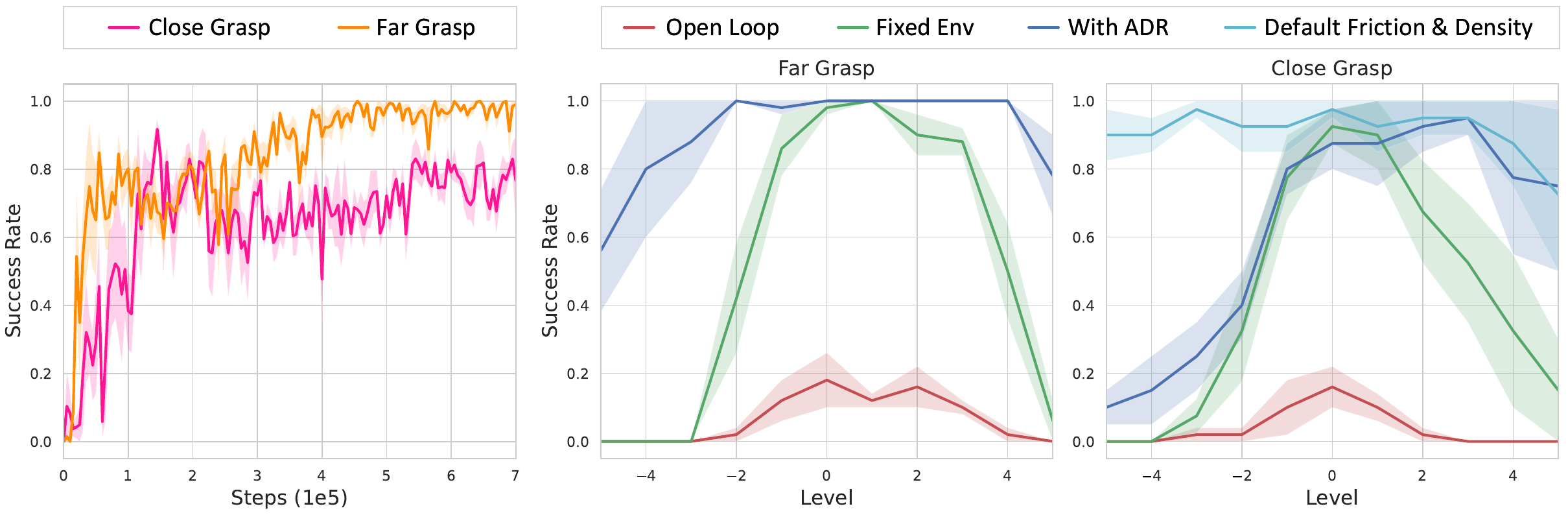}}
	\caption{
	\textbf{Left}: Training curves without ADR.
	\textbf{Right}: The performance of policies in different environments.
	A level with a larger absolute value represents a set of environmental parameters that are more different from the default parameters.
}
	\label{fig_generalization_env}
\end{figure}

\begin{figure}[tp]
	\centering
	{\includegraphics[scale=0.32]{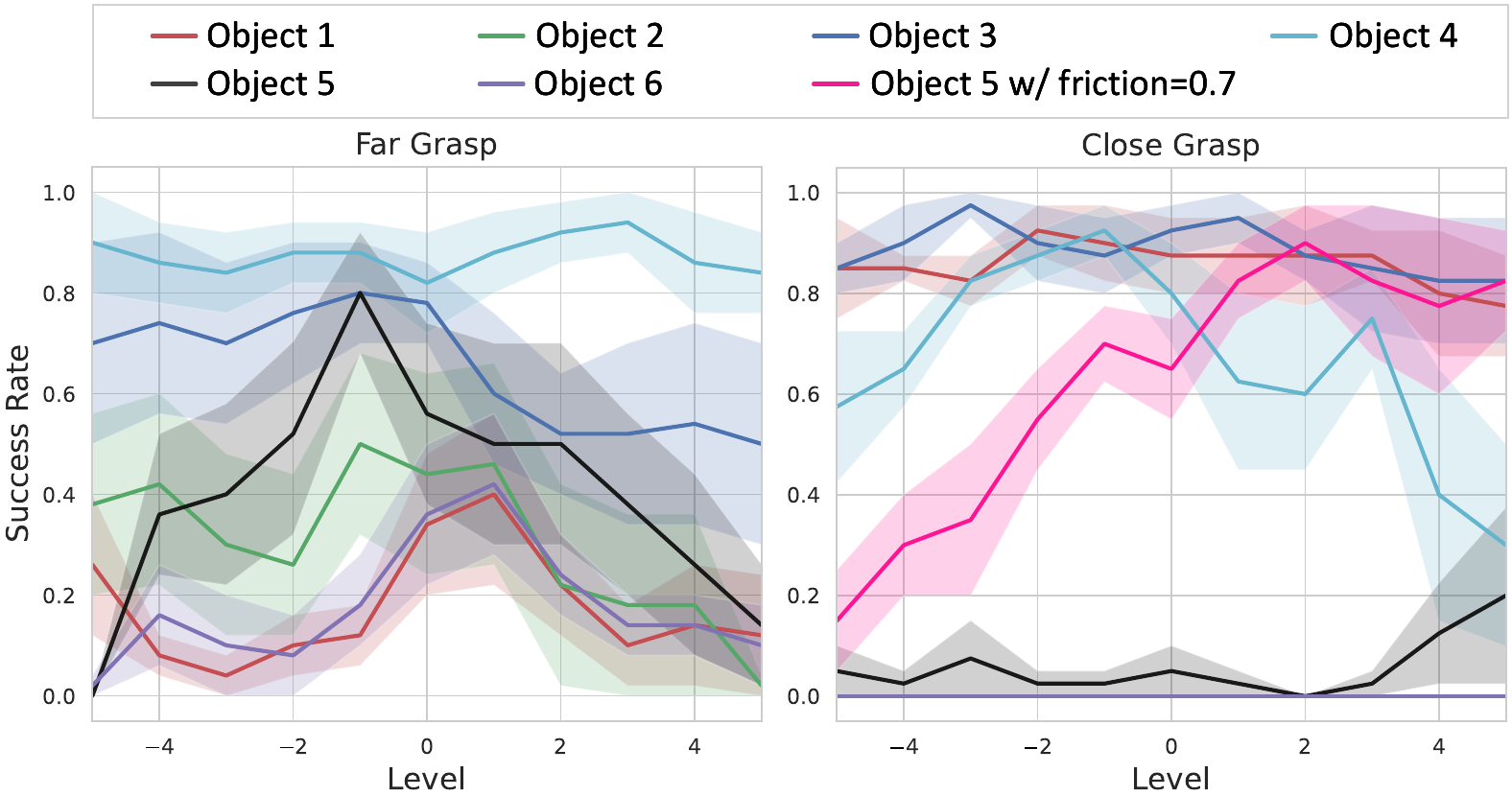}}
	\caption{
	The performance of MRLM with ADR on different objects.
}
	\label{fig_generalization_obj}
\end{figure}

\subsection{Policy Generalization to Environments}
\label{generalization_env}
In this section, we test the generalization of MRLM trained with different environmental parameters: open loop trajectories (\textit{Open Loop}), MRLM trained over a fixed environment (\textit{Fixed Env}) and MRLM trained with ADR (\textit{With ADR}).
The open loop trajectories are obtained by rolling out the Fixed Env policies in the default environment. 
Fig.~\ref{fig_generalization_env} (a) shows the training curves of MRLM trained with ADR based on the pretrained models that are trained in the default environment.
We select 11 equally spaced sets of environment parameters from the range covered by ADR, and test the accuracy of each policy running 10 episodes in each environment (Fig.~\ref{fig_generalization_env} (b)).
Level 0 represents parameters for the default environment.

The success rate of the open-loop trajectories do not exceed 30\% at the highest.
The closed-loop policies trained over a fixed environment perform much better, but the success rate drops off sharply as the parameters approach the range boundaries.
With ADR, the generalization can be improved even further.
With close grasp, policies with ADR does not perform well in environments near the lower bound, because the object may bounce away during flipping due to the low density and friction.
We set the object density and friction coefficient of all environments to default values and retest the policy trained with ADR, and the success rate has increased by a maximum of 90\%.

\subsection{Policy Generalization to Object Shape}
\begin{figure}[tp]
	\centering
	{\includegraphics[scale=0.4]{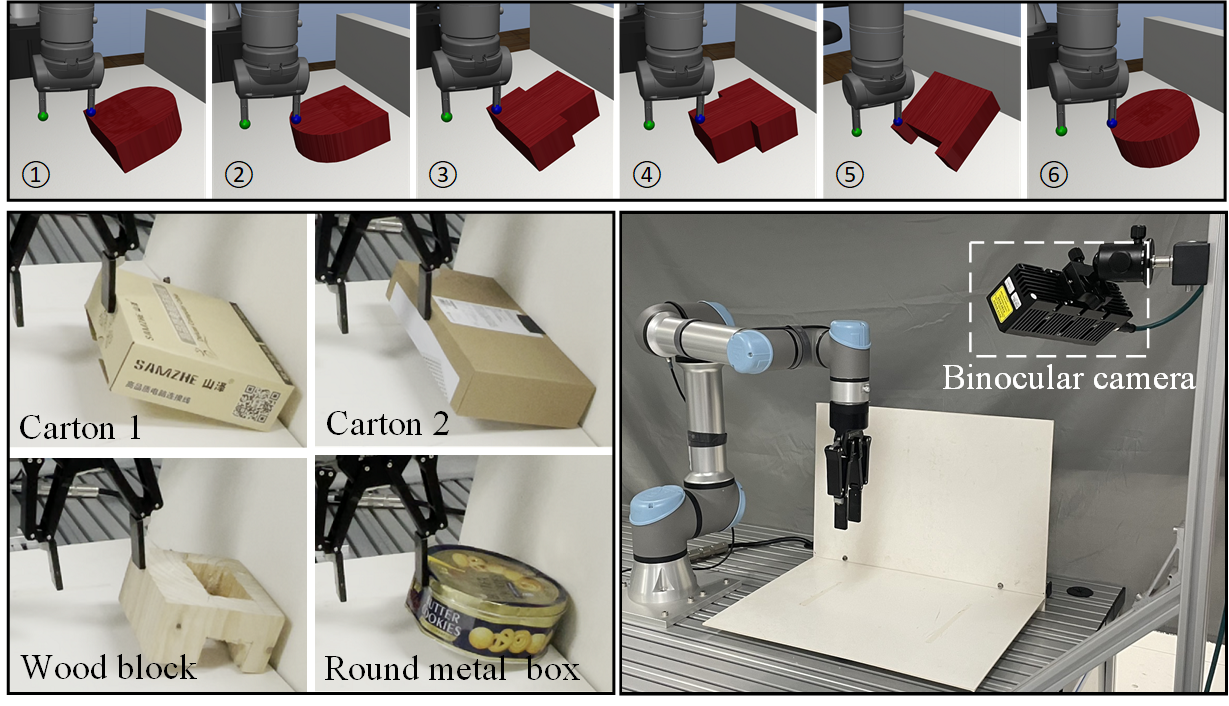}}
	\caption{
		\textbf{Top}: Test objects with different shapes in simulation environment.
		We use these objects to test MRLM trained only using boxes.
		\textbf{Bottom left}: Test objects in real world.
		\textbf{Bottom right}: Set-up for real world experiments.
	}
	\label{fig_setup}
\end{figure}

In this section, we test the generalization of MRLM to objects of different shapes outside the training distribution as shown in Fig.~\ref{fig_setup}.
In order to avoid object properties other than shape from interfering with the results, we set both the friction coefficient and density of the object to default values, and then test the success rate of the MRLM trained with ADR on each object in 11 environments described in Section.~\ref{generalization_env}.
The result is shown in Fig.~\ref{fig_generalization_obj}.

Object \ding{174} and \ding{175} have the highest success rates with the least oscillations because they are the closest in shape to boxes.
Objects whose contact surface with the table and wall is a curved surface have a lower success rate, with a maximum of 50\% and a minimum of 0\%.
We observe that the gripper has difficulty flipping the object \ding{176} with the default friction coefficient due to the offset gravity center.
We additionally test the success rate of object \ding{176} with close grasp with the friction coefficient set to 0.7, and the result is significantly improved, with an average increase of 60\%.
The hollow structure of the object \ding{176} makes it necessary to press the protruding parts or push the concave part to complete the flipping and tilting operations, which is difficult for the method of representing the object as a box.

\subsection{Real-robot experiments}
To perform real world experiments, we use a UR5e robot with a robotiq 2f-140 gripper.
We use a binocular camera fixed in front of the robot to obtain scene point clouds and object poses through point cloud registration.
We execute the MRLM on the robot with zero-shot sim2real transfer over 4 test objects with different shapes, densities, surface frictions, and sizes as shown in Fig.~\ref{fig_setup}.
We model objects before testing and obtain object poses during testing by running Principle Component Analysis (PCA) on point clouds.
Objects are placed in the scene according to position parameters randomly sampled in the range covered by ADR.
More details of the real robot experiments can be found in \href{https://sites.google.com/view/mrlm}{https://sites.google.com/view/mrlm}.
We evaluate 10 episodes for each test case and summarize the results in Table.~\ref{tab_real_result}. 
The success is measured by being able to close the gripper and lift the object at the end of the episode. 

Quantitatively, MRLM achieves a success rate of 95\% with close grasp while 67.5\% with far grasp.
Interestingly, MRLM achieves 9/10 successes over the round metal box with close grasp.
This demonstrates that although the policy is only trained with boxes in simulation, it can also generalize to other shapes to some extent. 
However, with far grasp, MRLM could fail to tilt the object and grasp it.

Qualitatively, MRLM exhibits similar strategies as discussed in Section 5.2. 
Similar to experiments in simulation with close grasp, MRLM demonstrates the ability to recover from failed flipping.
But the ability to recover from failed tilting transfers little from simulation to real scenarios.
We speculate that it is caused by the deviation between the real scene and the simulated scene.
The failure cases with close grasp are caused by the object being pushed away during the flipping.
The failure cases with far grasp are mainly because the object deviates from the intermediate pose in stage 2 and falls while being tilting.

\begin{table}[tp]
	\caption{Environment parameters in the experiment.}
	\footnotesize 
	\begin{center}
		\resizebox{\linewidth}{!}{
		\begin{threeparttable}
			\begin{tabular}{l|ccccc}
				\toprule[1pt]
				
				Object & Size (cm) & \makecell[c]{Surface\\Material} & Weight (g) & \makecell[c]{Success \\ w/ clo.} & \makecell[c]{Success \\ w/ far.} \\
				
				\midrule[1pt]
				
				Carton 1 & 18x18x6 & Cardboard & 95 			& 9/10 & 7/10 \\
				Carton 2 & 29.2x15.4x5.8 & Cardboard & 130 		& 10/10 & 6/10 \\
				Wood block & 14.5x14.5x8 & Wood & 645 			& 10/10 & 9/10\\
				Round metal box & 18.6x18.6x7.4 & Aluminum & 115 & 9/10 & 5/10\\
				\bottomrule[1pt]
				Average &&&& 95\% & 67.5\% \\
				\bottomrule[1pt]
			\end{tabular}
			\label{tab_real_result}
		\end{threeparttable}
	}
	\begin{tablenotes}
		\footnotesize
		\item[*] 
		\textbf{clo.:} Close grasp.
		\textbf{far.:} Far grasp.
	\end{tablenotes}
	\end{center}
\end{table}

\section{Discussion}
MRLM divides the long-horizon non-prehensile manipulation task into multiple stages to learn sequentially.
Compared to single-stage methods \cite{zhou2023learning1}, decomposing tasks according to object pose and contact point makes it possible to extend MRLM to more complex tasks involving multiple objects.
Compared with methods based on primitive skill combinations \cite{son2021reinforcement, shankar2019discovering}, MRLM enables robots to perform dynamically complex operations beyond primitive skills, such as flipping and tilting.

The limitations of MRLM include the limited number and position of intermediate object poses and the time-consuming training process.
It is possible to improve the quality of intermediate poses by first generating rough intermediate poses through a deep learning model and then refining them based on environment parameters.
The training process of MRLM in the default environment takes about 30 hours on our device with a RTX 3090Ti GPU and Intel i9-12900K CPU.
Learning from demonstrations and generalizing to new tasks may improve learning efficiency.









\bibliographystyle{Bibliography/IEEEtranTIE}
\bibliography{Bibliography/IEEEabrv, Bibliography/reference}\

\begin{thebibliography}{10}
\providecommand{\url}[1]{#1}
\csname url@samestyle\endcsname
\providecommand{\newblock}{\relax}
\providecommand{\bibinfo}[2]{#2}
\providecommand{\BIBentrySTDinterwordspacing}{\spaceskip=0pt\relax}
\providecommand{\BIBentryALTinterwordstretchfactor}{4}
\providecommand{\BIBentryALTinterwordspacing}{\spaceskip=\fontdimen2\font plus
\BIBentryALTinterwordstretchfactor\fontdimen3\font minus
  \fontdimen4\font\relax}
\providecommand{\BIBforeignlanguage}[2]{{%
\expandafter\ifx\csname l@#1\endcsname\relax
\typeout{** WARNING: IEEEtran.bst: No hyphenation pattern has been}%
\typeout{** loaded for the language `#1'. Using the pattern for}%
\typeout{** the default language instead.}%
\else
\language=\csname l@#1\endcsname
\fi
#2}}
\providecommand{\BIBdecl}{\relax}
\BIBdecl

\bibitem{pollayil2021planning}
G.~J. Pollayil, G.~Grioli, M.~Bonilla, and A.~Bicchi, ``Planning robotic
  manipulation with tight environment constraints,'' in \emph{2021 IEEE/RSJ
  International Conference on Intelligent Robots and Systems (IROS)}, pp.
  9385--9392.\hskip 1em plus 0.5em minus 0.4em\relax IEEE, 2021.

\bibitem{jiang2022path}
L.~Jiang, S.~Liu, Y.~Cui, and H.~Jiang, ``Path planning for robotic manipulator
  in complex multi-obstacle environment based on improved\_rrt,''
  \emph{IEEE/ASME transactions on mechatronics}, vol.~27, no.~6, pp.
  4774--4785, 2022.

\bibitem{liang2022search}
J.~Liang, M.~Sharma, A.~LaGrassa, S.~Vats, S.~Saxena, and O.~Kroemer,
  ``Search-based task planning with learned skill effect models for lifelong
  robotic manipulation,'' in \emph{2022 International Conference on Robotics
  and Automation (ICRA)}, pp. 6351--6357.\hskip 1em plus 0.5em minus
  0.4em\relax IEEE, 2022.

\bibitem{garrett2015backward}
C.~R. Garrett, T.~Lozano-P{\'e}rez, and L.~P. Kaelbling, ``Backward-forward
  search for manipulation planning,'' in \emph{2015 IEEE/RSJ International
  Conference on Intelligent Robots and Systems (IROS)}, pp. 6366--6373.\hskip
  1em plus 0.5em minus 0.4em\relax IEEE, 2015.

\bibitem{junge2020improving}
K.~Junge, J.~Hughes, T.~G. Thuruthel, and F.~Iida, ``Improving robotic cooking
  using batch bayesian optimization,'' \emph{IEEE Robotics and Automation
  Letters}, vol.~5, no.~2, pp. 760--765, 2020.

\bibitem{stouraitis2020online}
T.~Stouraitis, I.~Chatzinikolaidis, M.~Gienger, and S.~Vijayakumar, ``Online
  hybrid motion planning for dyadic collaborative manipulation via bilevel
  optimization,'' \emph{IEEE Transactions on Robotics}, vol.~36, no.~5, pp.
  1452--1471, 2020.

\bibitem{sun2020learning}
Z.~Sun, K.~Yuan, W.~Hu, C.~Yang, and Z.~Li, ``Learning pregrasp manipulation of
  objects from ungraspable poses,'' in \emph{2020 IEEE International Conference
  on Robotics and Automation (ICRA)}, pp. 9917--9923.\hskip 1em plus 0.5em
  minus 0.4em\relax IEEE, 2020.

\bibitem{zhou2023learning1}
W.~Zhou and D.~Held, ``Learning to grasp the ungraspable with emergent
  extrinsic dexterity,'' in \emph{Conference on Robot Learning}, pp.
  150--160.\hskip 1em plus 0.5em minus 0.4em\relax PMLR, 2023.

\bibitem{zhou2023learning2}
W.~Zhou, B.~Jiang, F.~Yang, C.~Paxton, and D.~Held, ``Learning hybrid
  actor-critic maps for 6d non-prehensile manipulation,'' \emph{arXiv preprint
  arXiv:2305.03942}, 2023.

\bibitem{kroemer2021review}
O.~Kroemer, S.~Niekum, and G.~Konidaris, ``A review of robot learning for
  manipulation: Challenges, representations, and algorithms,'' \emph{The
  Journal of Machine Learning Research}, vol.~22, no.~1, pp. 1395--1476, 2021.

\bibitem{gao2022fast}
K.~Gao, D.~Lau, B.~Huang, K.~E. Bekris, and J.~Yu, ``Fast high-quality tabletop
  rearrangement in bounded workspace,'' in \emph{2022 International Conference
  on Robotics and Automation (ICRA)}, pp. 1961--1967.\hskip 1em plus 0.5em
  minus 0.4em\relax IEEE, 2022.

\bibitem{shridhar2022cliport}
M.~Shridhar, L.~Manuelli, and D.~Fox, ``Cliport: What and where pathways for
  robotic manipulation,'' in \emph{Conference on Robot Learning}, pp.
  894--906.\hskip 1em plus 0.5em minus 0.4em\relax PMLR, 2022.

\bibitem{cheng2022contact}
X.~Cheng, E.~Huang, Y.~Hou, and M.~T. Mason, ``Contact mode guided motion
  planning for quasidynamic dexterous manipulation in 3d,'' in \emph{2022
  International Conference on Robotics and Automation (ICRA)}, pp.
  2730--2736.\hskip 1em plus 0.5em minus 0.4em\relax IEEE, 2022.

\bibitem{yang2021hierarchical}
X.~Yang, Z.~Ji, J.~Wu, Y.-K. Lai, C.~Wei, G.~Liu, and R.~Setchi, ``Hierarchical
  reinforcement learning with universal policies for multistep robotic
  manipulation,'' \emph{IEEE Transactions on Neural Networks and Learning
  Systems}, vol.~33, no.~9, pp. 4727--4741, 2021.

\bibitem{jang2005visibility}
H.-Y. Jang, H.~Moradi, S.~Lee, and J.~Han, ``A visibility-based accessibility
  analysis of the grasp points for real-time manipulation,'' in \emph{2005
  IEEE/RSJ International Conference on Intelligent Robots and Systems}, pp.
  3111--3116.\hskip 1em plus 0.5em minus 0.4em\relax IEEE, 2005.

\bibitem{rusu2009close}
R.~B. Rusu, N.~Blodow, Z.~C. Marton, and M.~Beetz, ``Close-range scene
  segmentation and reconstruction of 3d point cloud maps for mobile
  manipulation in domestic environments,'' in \emph{2009 IEEE/RSJ International
  Conference on Intelligent Robots and Systems}, pp. 1--6.\hskip 1em plus 0.5em
  minus 0.4em\relax IEEE, 2009.

\bibitem{hoang2022voting}
D.-C. Hoang, J.~A. Stork, and T.~Stoyanov, ``Voting and attention-based pose
  relation learning for object pose estimation from 3d point clouds,''
  \emph{IEEE Robotics and Automation Letters}, vol.~7, no.~4, pp. 8980--8987,
  2022.

\bibitem{wang2022six}
R.~Wang, C.~Su, H.~Yu, and S.~Wang, ``Six-dimensional target pose estimation
  for robot autonomous manipulation: Methodology and verification,'' \emph{IEEE
  Transactions on Cognitive and Developmental Systems}, vol.~15, no.~1, pp.
  186--197, 2022.

\bibitem{tang2022track}
T.~Tang and M.~Tomizuka, ``Track deformable objects from point clouds with
  structure preserved registration,'' \emph{The International Journal of
  Robotics Research}, vol.~41, no.~6, pp. 599--614, 2022.

\bibitem{huang2022task}
J.~Huang and K.~S. Au, ``Task-oriented grasping position selection in
  deformable object manipulation,'' \emph{IEEE Robotics and Automation
  Letters}, vol.~8, no.~2, pp. 776--783, 2022.

\bibitem{lu2022excavation}
Q.~Lu, Y.~Zhu, and L.~Zhang, ``Excavation reinforcement learning using
  geometric representation,'' \emph{IEEE Robotics and Automation Letters},
  vol.~7, no.~2, pp. 4472--4479, 2022.

\bibitem{you2021omnihang}
Y.~You, L.~Shao, T.~Migimatsu, and J.~Bohg, ``Omnihang: Learning to hang
  arbitrary objects using contact point correspondences and neural collision
  estimation,'' in \emph{2021 IEEE International Conference on Robotics and
  Automation (ICRA)}, pp. 5921--5927.\hskip 1em plus 0.5em minus 0.4em\relax
  IEEE, 2021.

\bibitem{qin2023dexpoint}
Y.~Qin, B.~Huang, Z.-H. Yin, H.~Su, and X.~Wang, ``Dexpoint: Generalizable
  point cloud reinforcement learning for sim-to-real dexterous manipulation,''
  in \emph{Conference on Robot Learning}, pp. 594--605.\hskip 1em plus 0.5em
  minus 0.4em\relax PMLR, 2023.

\bibitem{liu2022frame}
M.~Liu, X.~Li, Z.~Ling, Y.~Li, and H.~Su, ``Frame mining: a free lunch for
  learning robotic manipulation from 3d point clouds,'' \emph{arXiv preprint
  arXiv:2210.07442}, 2022.

\bibitem{qi2017pointnet}
C.~R. Qi, H.~Su, K.~Mo, and L.~J. Guibas, ``Pointnet: Deep learning on point
  sets for 3d classification and segmentation,'' in \emph{Proceedings of the
  IEEE conference on computer vision and pattern recognition}, pp. 652--660,
  2017.

\bibitem{yang2016survey}
L.~Yang, J.~Qi, D.~Song, J.~Xiao, J.~Han, Y.~Xia \emph{et~al.}, ``Survey of
  robot 3d path planning algorithms,'' \emph{Journal of Control Science and
  Engineering}, vol. 2016, 2016.

\bibitem{akkaya2019solving}
I.~Akkaya, M.~Andrychowicz, M.~Chociej, M.~Litwin, B.~McGrew, A.~Petron,
  A.~Paino, M.~Plappert, G.~Powell, R.~Ribas \emph{et~al.}, ``Solving rubik's
  cube with a robot hand,'' \emph{arXiv preprint arXiv:1910.07113}, 2019.

\bibitem{robosuite2020}
Y.~Zhu, J.~Wong, A.~Mandlekar, R.~Mart\'{i}n-Mart\'{i}n, A.~Joshi,
  S.~Nasiriany, and Y.~Zhu, ``robosuite: A modular simulation framework and
  benchmark for robot learning,'' in \emph{arXiv preprint arXiv:2009.12293},
  2020.

\bibitem{todorov2012mujoco}
E.~Todorov, T.~Erez, and Y.~Tassa, ``Mujoco: A physics engine for model-based
  control,'' in \emph{2012 IEEE/RSJ International Conference on Intelligent
  Robots and Systems}, \href{http://dx.doi.org/10.1109/IROS.2012.6386109}{DOI
  10.1109/IROS.2012.6386109}, pp. 5026--5033.\hskip 1em plus 0.5em minus
  0.4em\relax IEEE, 2012.

\bibitem{son2021reinforcement}
D.~Son, M.~Kim, J.~Sim, and W.~Shin, ``Reinforcement learning for vision-based
  object manipulation with non-parametric policy and action primitives,'' in
  \emph{2021 IEEE/RSJ International Conference on Intelligent Robots and
  Systems (IROS)}, pp. 5756--5763.\hskip 1em plus 0.5em minus 0.4em\relax IEEE,
  2021.

\bibitem{shankar2019discovering}
T.~Shankar, S.~Tulsiani, L.~Pinto, and A.~Gupta, ``Discovering motor programs
  by recomposing demonstrations,'' in \emph{International Conference on
  Learning Representations}, 2019.

\end{thebibliography}

\end{document}